%% file: main.tex
\documentclass[letterpaper, 10 pt, conference]{ieeeconf}  
\IEEEoverridecommandlockouts
\usepackage{graphicx}
\usepackage{hyperref}
\usepackage[nolist,nohyperlinks]{acronym} 
\usepackage{siunitx} 
\usepackage{amsmath,amsfonts,amssymb} 
\usepackage{subfigure} 
\usepackage[utf8]{inputenc} 
\usepackage[T1]{fontenc} 
\usepackage{caption}
\usepackage{capt-of}

\graphicspath{{images/}{../images}{../attachements}{attachements/}} 
\renewcommand\vec{\mathbf} 
\newcommand{\mat}{\mathbf} 

\title{Ascento: A Two-Wheeled Jumping Robot}

\author{Victor~Klemm$^*$,~Alessandro~Morra$^*$,~Ciro~Salzmann$^*$,~Florian~Tschopp,~Karen~Bodie,\\
~Lionel~Gulich,~Nicola~Küng,~Dominik~Mannhart,~Corentin~Pfister,~Marcus~Vierneisel,\\
~Florian~Weber,~Robin~Deuber, and Roland~Siegwart
\thanks{$^*$Contributed equally to this work.}
\thanks{All authors are members of the Autonomous Systems Lab, ETH Zurich, Switzerland. Contact: vklemm@ethz.ch, morraa@ethz.ch, sciro@ethz.ch}
}

\begin{document}

\maketitle
\thispagestyle{empty}
\pagestyle{empty}

\acrodef{ASL}[ASL]{Autonomous Systems Lab}
\acrodef{IMU}[IMU]{inertial measurement unit}
\acrodef{ToF}[ToF]{time-of-flight}
\acrodef{PA12}[PA12]{polyamide 12}
\acrodef{SLS}[SLS]{selective laser sintering}
\acrodef{CAD}[CAD]{computer-aided design}
\acrodef{EC}[EC]{electronically commutated}
\acrodef{DC}[DC]{direct current}
\acrodef{CAN}[CAN]{controller area network}
\acrodef{PID}[PID]{proportional integral derivative}
\acrodef{GUI}[GUI]{graphical user interface}
\acrodef{LiPo}[LiPo]{lithium-ion polymer}
\acrodef{ROS}[ROS]{robot operating system}
\acrodef{LQR}[LQR]{linear quadratic regulator}
\acrodef{MPC}[MPC]{model predictive control}
\acrodef{FEM}[FEM]{finite element method}


\begin{abstract}
Applications of mobile ground robots demand high speed and agility while navigating in complex indoor environments. 
These present an ongoing challenge in mobile robotics. 
A system with these specifications would be of great use for a wide range of indoor inspection tasks.
This paper introduces \textit{Ascento}, a compact wheeled bipedal robot that is able to move quickly on flat terrain, and to overcome obstacles by jumping.
The mechanical design and overall architecture of the system is presented, as well as the development of various controllers for different scenarios.
A series of experiments\footnote{Video accompanying paper: \url{https://youtu.be/U8bIsUPX1ZU}} with the final prototype system validate these behaviors in realistic scenarios. 
\end{abstract}

\section{Introduction}\label{sec:introduction}
    \input{chapters/1.introduction.tex}

\section{System Description}\label{sec:system}
    \input{chapters/2.system_description.tex}
    
\section{Modeling}\label{sec:modeling}
    \input{chapters/3.model.tex}
    
\section{Control} \label{sec:control}
    \input{chapters/4.control.tex}

\section{Experiments} \label{sec:experiments}
    \input{chapters/5.evaluation.tex}
    
\section{Experimental Features} \label{sec:features}
    \input{chapters/6.features.tex}

\section{Conclusion} \label{sec:conclusion}
    \input{chapters/7.conclusion.tex}

\section*{Acknowledgment}
The authors would like to acknowledge the generous support of ETH Zurich and the \ac{ASL}, Swiss Robotics, ANYbotics, Thyssenkrupp, Anewa AG, Maxon Motors, Wyss Zurich, Conrad, Matrix Vision and pmdtechnologies for their financial, technical, and moral support, without which this project would not have been possible.

\bibliographystyle{IEEEtran}
\bibliography{bibliography.bib} 

\end{document}

%% file: chapters/1.introduction.tex
The emergence of competent mobile robots over the past decade has pushed the development of inspection robotics in research and industry \cite{Hutter2017}, \cite{illahautonomous}.
While flying systems such as drones already show great maneuverability \cite{DBLP:journals/corr/abs-1801-04581}, they are very limited in payload and flight time.
Ground robots that can navigate quickly and master indoor obstacles are still the topic of ongoing research, and typically lack speed or versatility in their maneuvers.

The locomotion of ground robots can be categorized roughly into two main fields; either a leg- and foot-based \cite{Hutter2017a}, \cite{spotmini}, \cite{atlas}, \cite{asimo}, \cite{Armour_2007} or a wheel-based approach \cite{vespa}, \cite{rezero}.
Whilst some walking robots for indoor spaces show great performance overcoming obstacles such as stairs or slippery terrain, they usually still take a significant amount of time to execute these complex movements.
In contrast, robots with rotating elements such as wheels are well suited for flat grounds as they can move smoothly, efficiently and fast.
However, they usually fail to handle rough terrain, especially if there are obstacles larger than their wheel's radius.
An exception are systems with continuous tracks \cite{garm}.
These can overcome rough terrain and small obstacles with ease but are imprecise and inefficient for turning maneuvers due to slippage.

\begin{figure}[ht]
    \centering
    \includegraphics[width=1\linewidth]{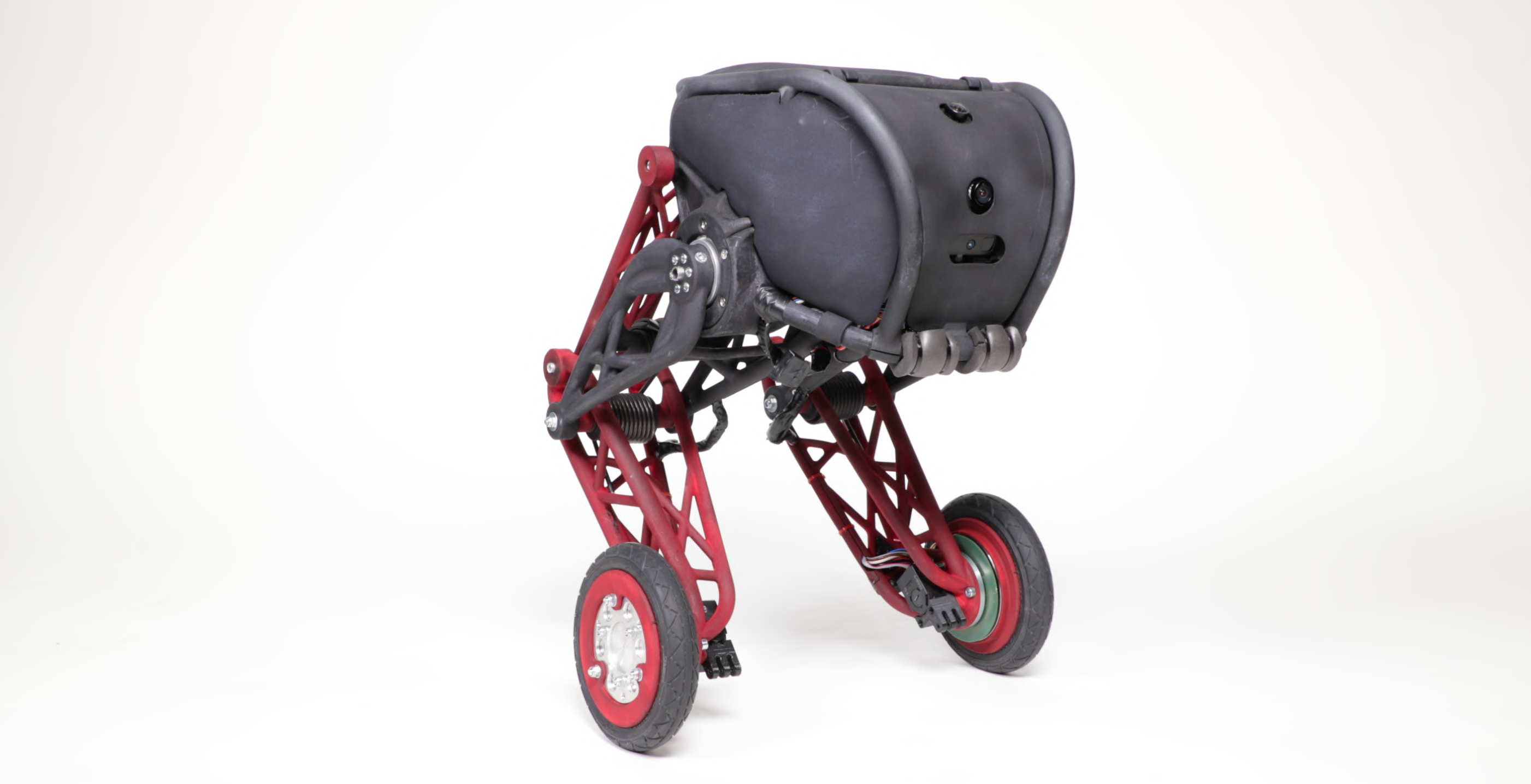}
    \caption{Current prototype of the \textit{Ascento} robot.}
    \label{fig:tina}
\end{figure}

Robots which combine these two core abilities, fast and smooth maneuvering on flat grounds and dynamic overcoming of obstacles, are rare.
Most systems are designed primarily for either one of those tasks, neglecting their performance on the other.

Our goal is to develop an indoor robot that combines the versatility of legs to overcome obstacles by jumping and wheels, which are efficient for fast movement on flat ground.
Tight indoor environments may compromise a robot’s mobility to a large extent.
Accordingly, the system must come in a highly compact design.

Current systems, which combine jumping and driving, exist and focus either on jumping very high at the cost of
repeatability \cite{urban_hopper}, \cite{doi:10.1177/1729881417745608} or require a significantly long time to recharge their jumping mechanism \cite{5373766}.
To the best knowledge of the authors, the system presented in \cite{handle} is the most similar design to the \textit{Ascento} robot.
It manages to combine legs and wheels efficiently for this purpose, but its size renders it unfit for indoor spaces. 
In this work, we present \textit{Ascento}, a two-wheeled jumping robot in a small form factor especially well-suited for indoor environments. 
The structural components were created with topology optimization and are fully 3D-printed.
An optimized leg geometry decouples the driving and jumping motion and allows the robot to recover from various fall scenarios.
Stabilization and driving is achieved through a \ac{LQR} controller. For jumping and fall recovery maneuvers, a sequential feed forward controller with feedback tracking is used.
To validate this concept, the real-world prototype demonstrated stabilized driving, jumping and fall recovery in multiple experiments.

The three main contributions can be summarized as:
\begin{itemize}
    \item The mechanical design of a two-wheeled balancing robot with a parallel elastic jumping mechanism, built from topology optimized parts.
    \item Dedicated software for controlling the robot.
    \item Successful experimental evaluation of the idea on a real world prototype.
\end{itemize}

The remainder of this paper is structured as follows:
In \autoref{sec:system}, we describe the system's mechanical design and list the integrated hardware.
In \autoref{sec:modeling}, a brief outline of the system's model used for model-based control is given.
An explanation of the system's control architecture is given in \autoref{sec:control}, highlighting the control strategies used for stabilizing \ref{subsec:stabilizingcontrol}, jumping \ref{subsec:jumpcontrol} and fall recovery \ref{subsec:standupcontrol}. The evaluation of real-world experiments is presented in \autoref{sec:experiments}.
We further list the robot's features which are under development in \autoref{sec:features}, before we conclude in \autoref{sec:conclusion}.

%% file: chapters/2.system_description.tex
\subsection{Mechanical Design}

\begin{figure}[ht]
    \centering
    \includegraphics[width=0.68\linewidth]{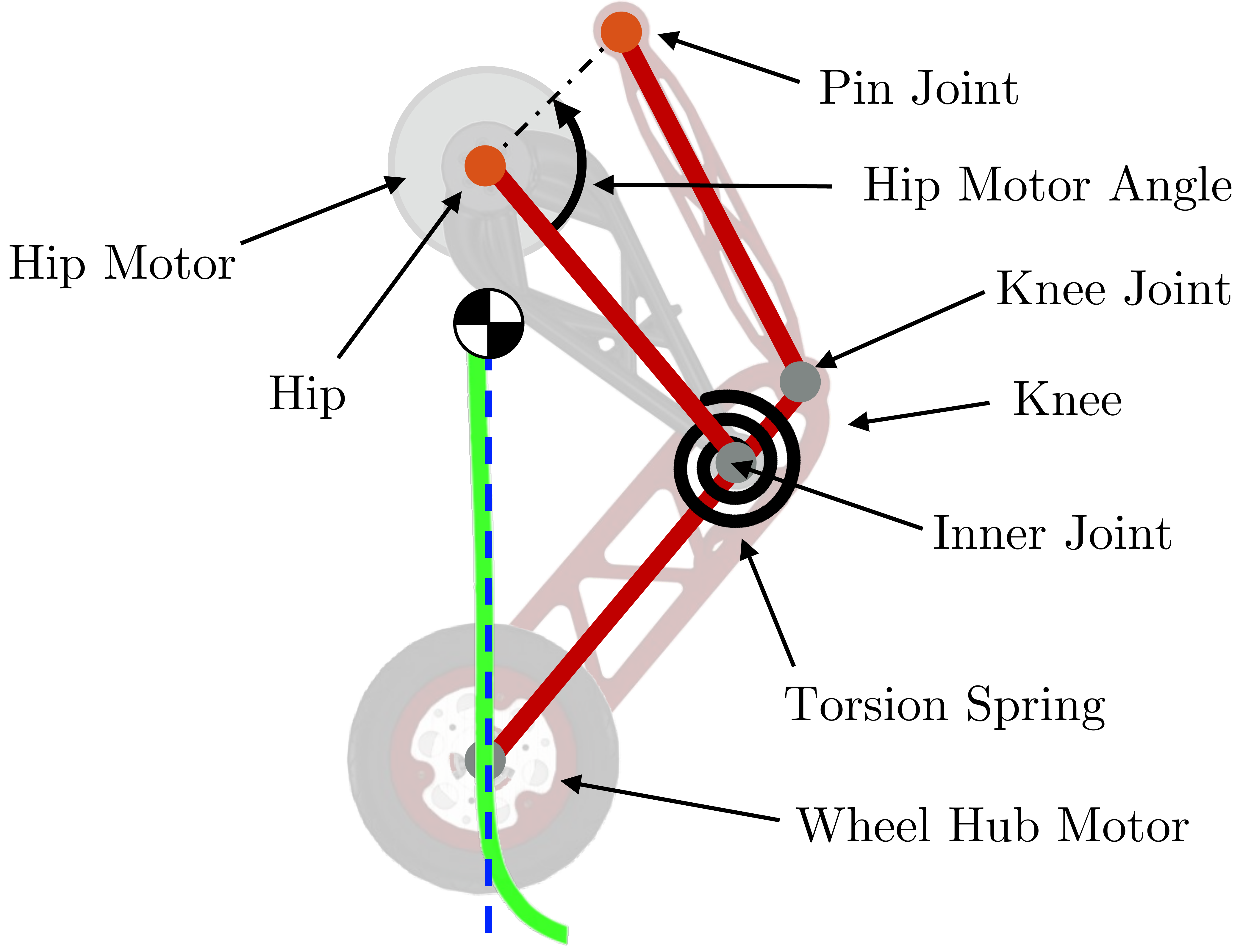}
    \caption{Left leg assembly and its main components. The optimal linear motion is depicted with a blue dotted line passing through the robot's center of mass and the achieved optimized motion is depicted in green.}
    \label{fig:leg_assembly}
\end{figure}

The presented system consists of two legs ending with actuated wheels. The legs are attached to a body that houses all electronics as shown in \autoref{fig:tina}.
Each leg can be extended and retracted independently by actuating the corresponding motor installed in the hip of the body. 
This way, the system's total height can be adjusted between \SI{31}{\cm} and \SI{66}{\cm}.
The goal of the leg mechanism is to decouple stabilizing and jumping control as much as possible.
This was realized through a three bar linkage which approximates a linear motion of the wheels perpendicular to the ground as seen in \autoref{fig:leg_assembly}.
By making the line pass through the system's center of mass the leg motion dictates a jump trajectory that does not cause a body rotation.
As such a mechanism cannot achieve a perfectly straight line \cite{hartenberg:kinematic_linkages}, the linkage is optimized numerically for six geometric design parameters with a mean squared error approach to the desired linear trajectory, whereby the optimal lengths and angles of the bars were obtained.

To reduce mass and increase the strength of the system the leg components have been designed using topology optimization \cite{Bends&oslash;e2001} inspired geometries. 
Topology optimization is a numerical method that finds a geometry for given set of loads, design space, boundary conditions and a target mass with the goal of minimizing stresses within the structure.
The raw topology optimized structure is impractical for post-processing steps and numerical analysis such as \ac{FEM}. 
The structure was therefore redesigned manually using the optimized structure as template.
Manufacturing the optimized parts would be challenging with classical manufacturing processes because of their complex shape.
Therefore, all structural parts are 3D-printed from \ac{PA12} using \ac{SLS} technology.
This technology has also enabled fast prototyping iterations.

To provide smooth stabilization and to counteract disturbances of the tilt angle of the system, near-zero backlash and high wheel torques are required.
For this purpose, a custom wheel assembly with frameless hub motors was constructed.
This direct drive configuration allows an almost backlash free motion and is very compact.
Torsion springs installed in the inner joints as shown in \autoref{fig:leg_assembly} counteract the system's own weight and reduce the control effort of the hip motors when driving or standing, increasing overall efficiency and jumping height.

\subsection{Hardware}

The remaining hardware components have been chosen to best fulfill the system's performance requirements.
To jump and be able to counteract the spring, when the system has no ground contact, high torques are needed in the hip motor.
For this purpose ANYdrive \cite{Hutter2017a} series-elastic actuators are used.
They can deliver high peak torques of up to \SI{40}{Nm} and have built-in position and torque control. 
Maxon EC90 frameless \ac{EC} motors with a maximum torque of \SI{3.5}{Nm} are utilized in the wheel hub motor.
Each wheel motor is equipped with an encoder for precise position and velocity feedback and requires an additional motion controller for torque control. 
All four motors communicate via a \ac{CAN} and use an adapter to communicate with the onboard computer. 
As the main processing unit, which takes full control of the system, an Intel NUC with i7 processor is used. 
The system is also equipped with an \ac{IMU} and a microcontroller to allow communication between the \ac{IMU} and the computer.
Two \ac{ToF} distance sensors used for triggering of a jump are installed next to the wheels.
To power all motors, a battery pack composed of four 3-cell \ac{LiPo} batteries connected in series is used. 
The onboard computer and the remaining electronic devices are powered by a single 4-cell \ac{LiPo} Battery. 
A complete component list is shown in \autoref{tab:data_used_hardware}.

As a whole, the presented system weights \SI{10.4}{\kg} and has an operation time of approximately \SI{1.5}{h}. 

\begin{table}[h]
\centering
\caption{Components and suppliers}
\begin{tabular}{|l | l|}
        \hline

        \textbf{Component} & \textbf{Name} \\ \hline
        
        Wheel Motor & Maxon EC90 Frameless\\\hline
        
        Wheel Motor Motion Controller & Maxon EPOS4 Compact\\\hline
        
        Hip Motor & ANYbotics ANYdrive \\\hline
        
        Wheel Motor USB-to-CAN & ixxat USB-to-CAN V2\\\hline
        
        Hip Motor USB-to-CAN & Lawicel CANUSB \\\hline
        
        Motor Battery & Hacker LiPo \SI{5000}{mAh} 3S\\\hline
        
        Computer Battery & Turnigy LiPo \SI{2200}{mAh} 4S\\\hline
        
        Onboard Computer & Intel NUC KIT NUC7i7BNH\\\hline
        
        Microcontroller & Arduino Uno\\\hline
        
        IMU & Analog Devices ADIS16460\\\hline
        
        Wheel Encoder & AEDL-5810-Z12\\\hline
        
        ToF Distance Sensors & Terabee TeraRanger Multiflex\\\hline
        
        3D Mouse & 3Dconnexion SpaceMouse\\\hline
        
        Gesture Control Device & Leap Motion Controller\\\hline
        
\end{tabular}
\label{tab:data_used_hardware}
\end{table}

\subsection{Software}\label{subsec:software}

\begin{figure}[ht]
    \centering
    \includegraphics[width=1.0\linewidth]{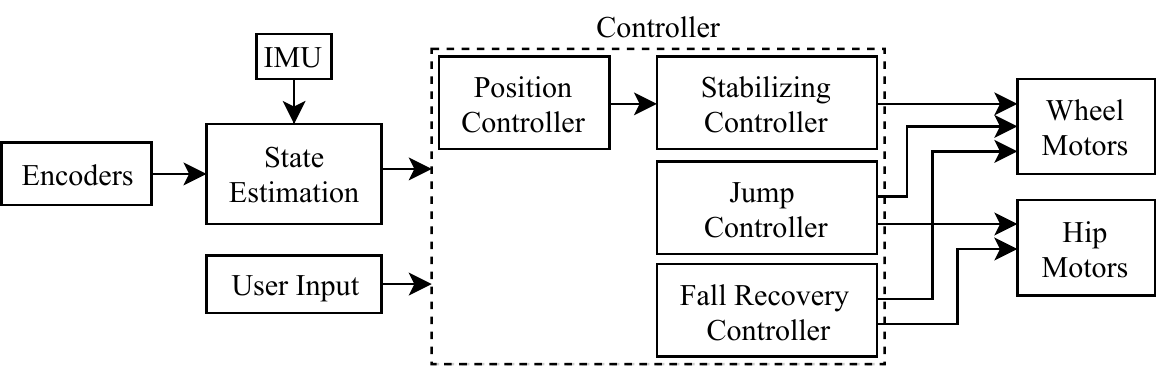}
    \caption{Overview of the controller architecture.}
    \label{fig:controller_architecture}
\end{figure}

The control-related software must be computationally efficient to enable high bandwidth controllers, hence all software is written using C++. 
In addition, the \ac{ROS} framework is used for high level communication.
A Kalman filter is implemented using sensor data obtained by the \ac{IMU} and motor encoder measurements.
In combination with the model knowledge from \ref{subsec:rbm}, the Kalman filter provides an estimate of the system's state dependent on the hip motor position as described in \ref{subsec:stabilizingcontrol}.
The estimated state information is fed to the controller together with the desired pose from the user as shown in \autoref{fig:controller_architecture}.
The user input originates either from a 3D mouse or a gesture control device \cite{Gulich2018} offering easy and intuitive steering.
The controller block includes the stabilizing, jump and fall recovery controllers as well as a high level position controller. 
Jumping and driving maneuvers are considered decoupled due to the optimized leg geometry and are therefore controlled independently.
The stabilizing controller computes and sends torque commands to actuate the wheel motors. 
Similarly, the jump and fall recovery controllers take full control of the hip and wheel motors.

%% file: chapters/3.model.tex
To apply model-based control strategies and advanced state estimation techniques a system model to describe the robot's rigid body dynamics is derived.

\subsection{Coordinates and Conventions}\label{subsec:coc}

\subsubsection{Notation convention}
$\dot{a}$ denotes a temporal derivative, $\hat{a}$ represents an estimate, $\mat{A}^{\intercal}$ a matrix transpose and $a_{l}$ a symmetrically occurring quantity on the left, $a_{r}$ on the right, respectively.

\subsubsection{Generalized coordinates and states}
\begin{figure}[ht]
    \centering
    \includegraphics[width=1\linewidth]{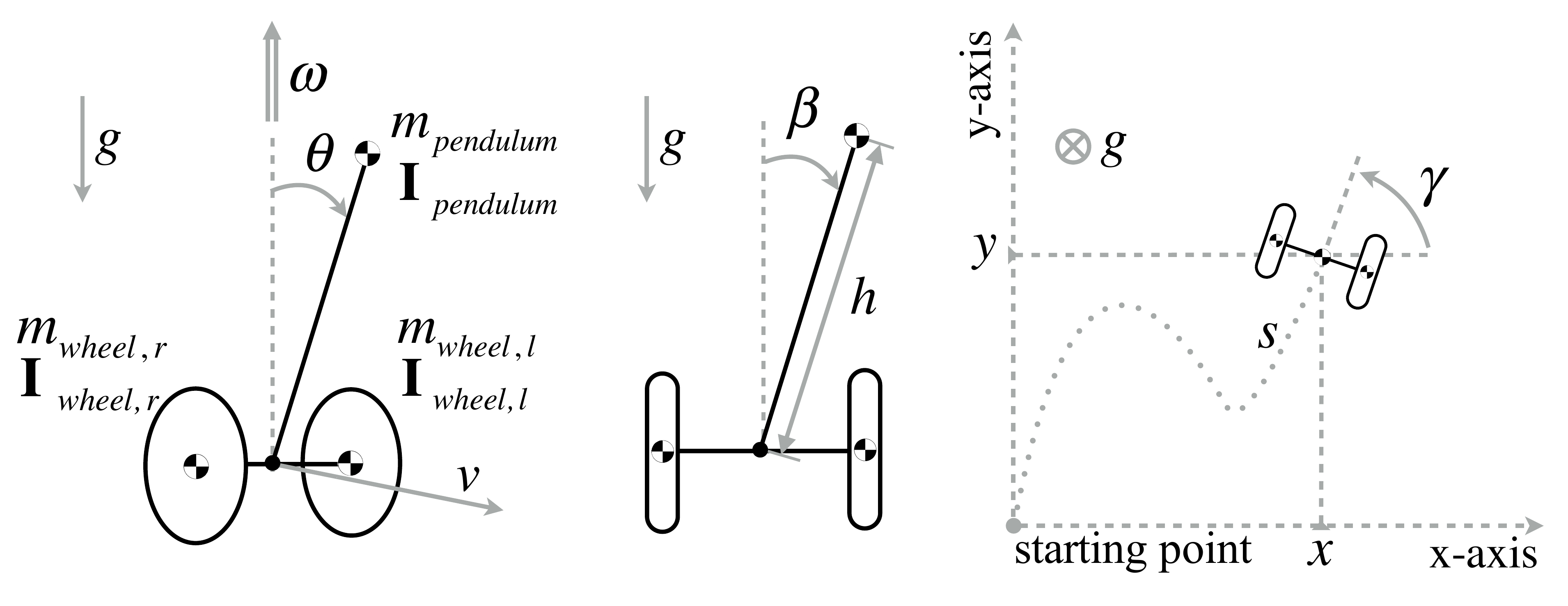}
    \caption{Generalized coordinates of the simplified \textit{Ascento} robot model, in a diagonal, back and top view. We use $m_{body}$ for mass and $\mat{I}_{body}$ for the inertia tensor with the corresponding name of the described body as a subscript. $g$ indicates the Earth's gravitational acceleration.}
    \label{fig:generalized_coordinates}
    \end{figure}

In order to model the simplified system as described in \ref{subsec:rbm}, we introduce a set of system state variables shown in \autoref{fig:generalized_coordinates}.
$\theta$ denotes the forward tilt angle of the robot, $v$ the planar linear velocity and $\omega$ the normal angular velocity.
The robot's leg positions are modeled by the sideways tilt angle $\beta$ (currently only used for an experimental lean mode introduced in \autoref{sec:features}) and the distance $h$ to the substitute center of mass.
For odometry considerations, we use $x$, $y$ and $\gamma$ as planar coordinates for position and orientation and $s$ as the traveled distance on the surface.

We further introduce the generalized coordinates vector $\vec{q}=[\theta \quad s \quad \gamma]^{\intercal}$, which is used in \ref{subsec:rbm} to model the system, and the \ac{LQR} state vector $\vec{x}=[\theta \quad \dot{\theta} \quad v \quad \omega]^{\intercal}$ for stabilization feedback control as described in \ref{subsec:stabilizingcontrol}.

\subsection{Assumptions}\label{ass}

To keep modeling efforts low and the model simple, the following assumptions were made:
\begin{enumerate}
    \item The dynamics of the leg linkages are neglected.
    \item Perfect joints with no  friction or hysteresis are assumed.
    \item Friction between floor and wheels is simplified by implying a no-slip condition.
    \item The motor dynamics are neglected, as they are significantly faster than the rest of the system.
    \item System delay is left unmodeled.
    \item All links and bodies are rigid.
\end{enumerate}
Assumption 1 implies a fixed hip motor, rendering the model only applicable to a specific leg configuration.
This limitation is addressed using interpolated control strategy which lowers the significance of the assumption as described in \autoref{sec:control}.
All remaining assumptions can successfully be modeled as unknown external disturbances and compensated by a sufficiently fast and robust controller as shown in \autoref{sec:experiments}.

\subsection{Rigid Body Model}\label{subsec:rbm}

The assumption of fixed leg geometry reduces the robot's model on a specific height to a two-wheeled inverted pendulum model, consisting of three bodies: Two wheels and an inverted pendulum body with a substitute mass, length and inertia tensor, combined from all included bodies.
Using spatial velocity-transport formulae \cite{mazza}, the kinematics of each body are formulated using only the generalized coordinates introduced in \ref{subsec:coc}.
From these expressions, the kinetic and potential energies of all bodies, $T$ and $V$, respectively, can be derived, leading to the Lagrangian energy function given by 
\begin{equation}
    \mathcal{L}=T-V.
\end{equation}
The equations of motion of the system are obtained by using the Lagrange equation of the first kind
\begin{equation}\label{lagrange}
\frac{d}{dt}(\frac{\partial \mathcal{L}}{\partial \dot{\vec{q}}})-\frac{\partial \mathcal{L}}{\partial \vec{q}} = \mat{J}^{\intercal}\cdot\vec{c}
\end{equation}
where $t$ is the continuous time variable, $\mat{J}$ is the Jacobian transformation matrix of the system and $\vec{c}$ is the external Cartesian forces vector.
In $\vec{c}$, the horizontal and vertical force components of all rigid bodies were set to zero.
The torque components were set equal to the particular motor and spring torques acting on the corresponding rigid body.

From \autoref{lagrange} the implicit equations of motion
\begin{equation}
\mat{M}\cdot\ddot{\vec{q}} = \vec{f}
\end{equation}
are derived, with $\mat{M}$ being the mass matrix of the system and $\vec{f}$ being the forcing term.

All system parameters, such as lengths, masses and moments of inertia, are determined from precise mass measurements of all used components and calculated by using the \ac{CAD} models, assuming constant density of all components.

%% file: chapters/4.control.tex
A two-wheeled robot is inherently unstable.
Thus, dedicated control strategies are required not only for jumping, but even standing still and driving. 
Additionally, to get into operational mode or recover from a fall a specific control maneuver is required.

\subsection{Stabilizing Control}
\label{subsec:stabilizingcontrol}
Being able to drive, jump and land again, while staying upright on two wheels, requires a reliable stabilization algorithm.
Robustness is also crucial. The robot should be able to handle external disturbances while using as little space as possible to regain its equilibrium.
The used approach is an \ac{LQR}, which is an optimal control strategy for regulating a linear system at minimal cost. 
Li, Yang and Fan \cite{Li:2012:ACW:2349034} showed that an LQR controller can successfully provide high reliability and robustness for two-wheeled inverted pendulum stabilization applications.

The optimal solution of the \ac{LQR}'s infinite horizon problem is found by solving the discrete-time algebraic Riccati equation
\begin{equation}
\mat{F}^{\intercal}\cdot\mat{S}\cdot\mat{G}\cdot[\mat{R}+\mat{G}^{\intercal}\cdot\mat{S}\cdot\mat{G}]^{-1}\mat{G}^{\intercal}\cdot\mat{S}\cdot\mat{F}+\mat{S}-\mat{F}^{\intercal}\cdot\mat{S}\cdot\mat{F}-\mat{Q}=\mat{0}
\end{equation}
with $\mat{F}$ and $\mat{G}$ being the discrete-time state-space representation matrices of the system, $\mat{Q}$ and $\mat{R}$ being the weight matrices and $\mat{S}$ being the unknown matrix of the equation.

The optimal feedback gain matrix $\mat{K}$ is given by
\begin{equation}
\mat{K} = [\mat{R}+\mat{G}^{\intercal}\cdot\mat{S}\cdot\mat{G}]^{-1}\cdot\mat{G}^{\intercal}\cdot\mat{S}\cdot\mat{F}. 
\end{equation}

$\mat{Q}$ and $\mat{R}$ were selected based on the importance of each state and the desired aggressiveness of the overall system.
To simplify this process, the weight matrices were assumed to be diagonal, which reduces the number of adjustable weight parameters to six.
To tune these values, intricate tests were performed on the real system.

The choice of the \ac{LQR} state vector as $\vec{x}=[\theta \quad \dot{\theta} \quad v \quad \omega]^{\intercal}$ omits spatial position and orientation.
The idea behind this selection is to leave the problem of position tracking to a dedicated controller which gives direct velocity commands to the \ac{LQR}.
When the operator steers the robot, a reference setpoint is added to $v$ or $\omega$ in the state vector before it is fed into the controller, thereby commanding the robot to reach a specific target velocity.

To take into account varying knee angles, the dynamic equations were linearized around ten different, equally spaced leg heights.
This yields ten fourth order state-space models and feedback gain matrices $\mat{K}$, between which linear interpolation is used.
Thereby, the restrictions of the simplified model introduced by assumption $1$ in \ref{ass} can be resolved.

The system is regulated by the \ac{LQR} control law
\begin{equation}
\vec{u}=-\mat{K}(\hat{h})\cdot\hat{\vec{x}}
\end{equation}
where the input vector $\vec{u}$ consists of the left and right wheel torque and $\hat{h}$ is used for linear interpolation between the gain matrices.
Here $\hat{\vec{x}}$ and $\hat{h}$ are directly supplied by the state observer, as described in \ref{subsec:software}.
According to the separation principle, this estimation and control setup is guaranteed to be stable and to lead to the robot returning to its perfectly upright equilibrium position as long as noise, disturbances, modeling errors and actuator saturation have no influence on the system's dynamics.

\subsection{Jump Control}
\label{subsec:jumpcontrol}
\begin{figure}[ht]
    \centering
    \includegraphics[width=1.0\linewidth]{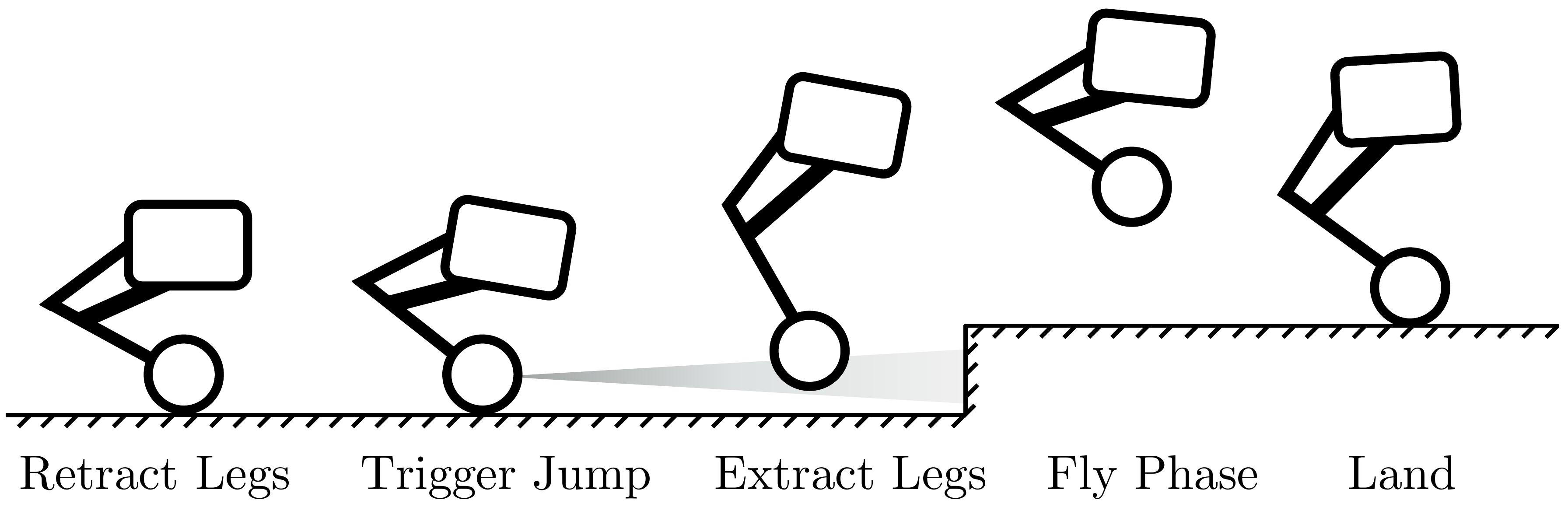}
    \caption{The discrete phases of the jump sequence. In each phase a different control strategy is applied.}
    \label{fig:jump_phases}
\end{figure}

The activation of the jump controller triggers a predefined jump sequence which overrides the current drive controllers and takes full control of the robot.
The jump controller is a heuristic feed-forward controller, inspired by human jumping motion, with discrete, successive phases (\autoref{fig:jump_phases}).
In the following, the jump phases required for the jump on a step are described in further detail.

\subsubsection{Retract Legs}
Using a controller that follows a specific trajectory for the hip motors, the robot's legs are retracted.
During this process, the stabilizing controller is active.

\subsubsection{Trigger Jump}
As soon as nominal stability after the height change is detected, the robot gathers forward velocity.
Using the \ac{ToF} distance sensors, the following leg extraction is triggered when a predefined distance to the step is reached.

\subsubsection{Extract Legs}
The legs are then extracted by the two hip motors, which are regulated by a \ac{PID} controller for synchronous extraction.
Once the legs are completely extracted, the stabilizing controller is disabled and ground contact is lost.

\subsubsection{Fly Phase}
A \ac{PID} loop is used on the hip motor positions with reference on a retracted leg position. Thereby, a virtual spring damper element is simulated.
This behaviour is desirable to either jump over high obstacles or prevent the wheels from touching the stair edge.

\subsubsection{Land}

Ground contact is detected when the torques in the hip joints exceed a specific threshold.
Upon detection of ground contact, stabilizing control is resumed.
Again, a virtual spring-damper element in the hip motors is simulated, allowing for a smooth energy dissipation and controlled landing.

The jump height and forward velocity during the jump can be set by the user via a \ac{GUI}. 
This adjusts the parameters of the jump phases for different scenarios such as jumping on spot, jumping while driving and jumping onto a step.

\subsection{Fall Recovery Control} \label{subsec:standupcontrol}
\begin{figure}[ht]
    \centering
    \includegraphics[width=0.9\linewidth]{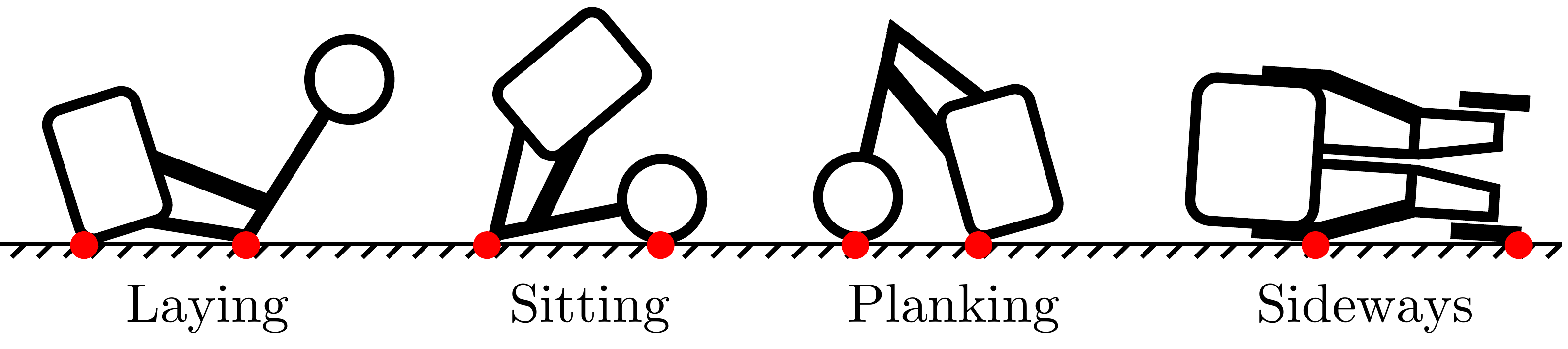}
    \caption{All four stable positions the robot may fall into. The contact points between robot and ground are represented by red dots.}
    \label{fig:standup_positions}
\end{figure}

After a fall or during start-up, the robot is not in its upright position. 
Stand up procedures are addressed for three out of four resting positions (\autoref{fig:standup_positions}) \cite{SalzmannWeber2018}.  
Furthermore, the system is able to go into these resting positions in a controlled manner.

The resting positions are defined by the contact points between ground and robot.
In the laying position, the robot contacts the ground with the legs and rear parts of the body. 
When in sitting position, it touches the ground with the legs and wheels and in the planking position with the front part of the body and wheels.
When the robot lays on its side contacting a single leg and wheel (sideways position), it is neither able to recover from nor achieve this resting position in a controlled manner.

\begin{figure}[ht]
    \centering
    \includegraphics[width=0.9\linewidth]{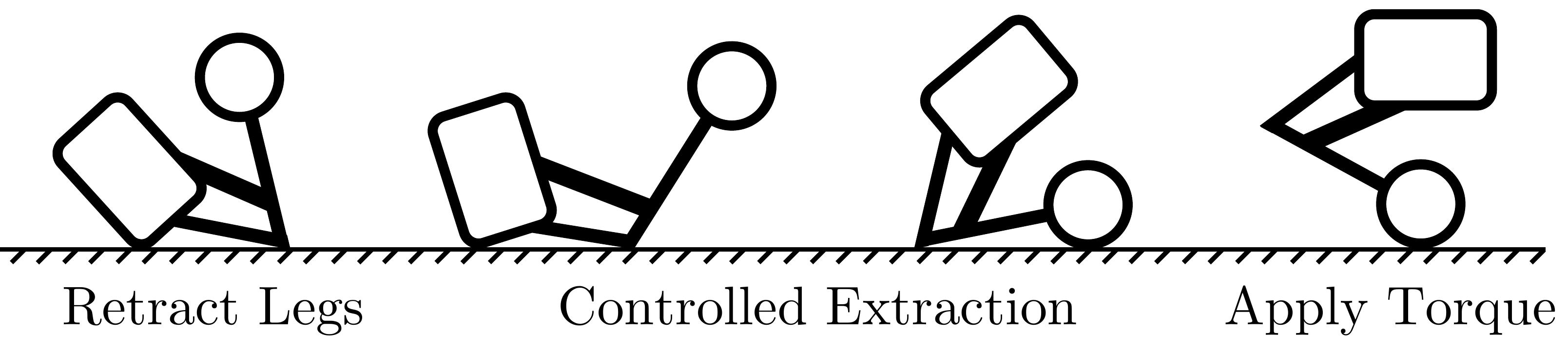}
    \caption{Phases of the stand up procedure. The controlled extraction step only applies for the laying position.}
    \label{fig:standup_procedure}
\end{figure}

Similar to the jump procedure, the stand up procedure is composed of discrete, successive phases (\autoref{fig:standup_procedure}) which are similar for all resting positions. 
The event is triggered by the user and overrides the current drive controller. 

\subsubsection{Retract Legs}
Using a controller that follows a specific trajectory for the hip motors, the robot's legs are retracted.

\subsubsection{Controlled Extraction}
This step only applies for the laying position. Following a predetermined trajectory, the legs are extracted and later retracted again. 
This induces a rotation around the knee, and the robot achieves the sitting position. 
A controlled extraction of the legs is required as too fast an extraction would make the robot lose ground contact and have a hard impact.
On the other hand, too slow an extraction would not suffice to tip the robot over to the sitting position.

\subsubsection{Apply Torque}
A constant torque is applied to the wheel motors, backwards for the sitting position and forwards for the planking position.
The constant torque is applied until the robot has enough rotational energy to reach zero tilt angle.
Once the robot stands vertically, the stabilizing controller is turned on and the robot brakes to reach its upright idle position.

Entering a resting position in a controlled manner is achieved by turning off the stabilizing controller and applying a small torque to the wheel motors to control the fall direction.

%% file: chapters/5.evaluation.tex
    \subsection{Simulation} 
    
    All the control algorithms were tested in a simulation, based on Gazebo \cite{gazebo} as a physics engine. 
    A model of the robot with mass and inertia values from the CAD model allows for a realistic and computationally efficient testing environment.
    The similar behaviour of the simulation and the experiments suggests the validity of \autoref{ass} and allows for further development \autoref{sec:features}.
   
    \subsection{Experimental Results}
    
\begin{figure}[ht]
    \parbox[t]{0.36\linewidth}{\null
        \centering
        \includegraphics[width=.36\columnwidth]{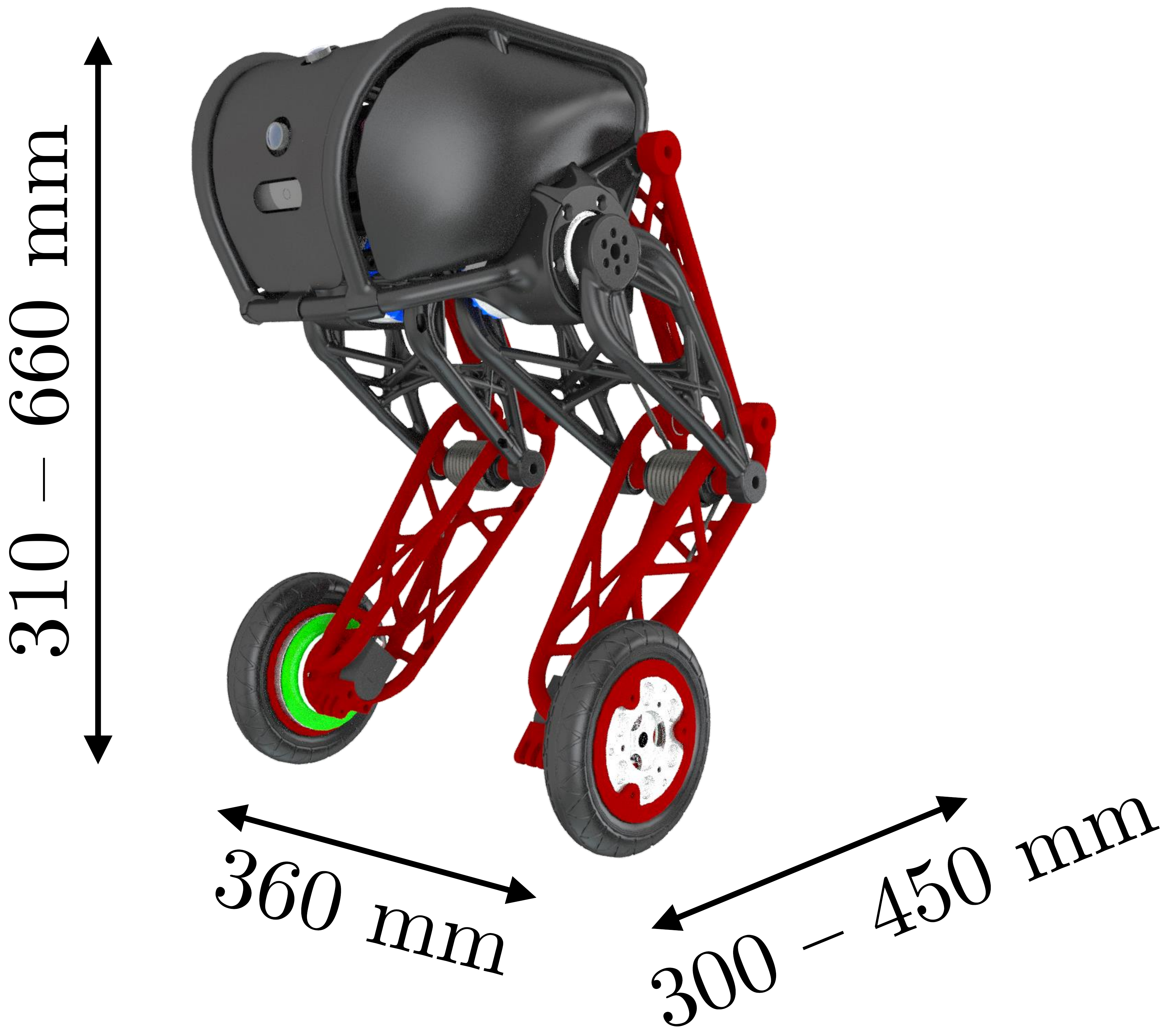}%
        \captionof{figure}{Main system dimensions.}%
        \label{fig:robot_dimensions}
    }
    \parbox[t]{0.60\linewidth}{\null
        \centering
        \vskip-\abovecaptionskip
        \captionof{table}[t]{Technical specifications of the robot.}%
        \label{tab:technical_specs}
        \vskip\abovecaptionskip
        \small
            \begin{tabular}{|l|l|}
            \hline
            \textbf{Category} & \textbf{Value} \\ \hline 
            Weight                   &  \SI{10.4}{\kilo\gram}      \\ \hline
            Max. linear velocity     &  \SI{8.0}{\kilo\meter\per\hour}     \\ \hline
            Max. angular velocity    &  \SI{1.1}{\radian\per\second}  \\ \hline
            Max. jumping height      &  \SI{0.4}{\meter}      \\ \hline
        Battery lifetime             &  \SI{1.5}{\hour}        \\ \hline
        \end{tabular}%
    }
    \end{figure}

    In this section, results from a series of experiments with the prototype system are presented. 
    The dimensions and other technical specifications of the robot are shown in \autoref{fig:robot_dimensions} and Table \ref{tab:technical_specs}, respectively.
    Multiple experiments are presented to demonstrate the robot's specific capabilities regarding stabilizing performance (\ref{subsec:stabilizingperformance}), jumping (\ref{subsec:jumpingevalu}) and fall recovery (\ref{subsec:standupprocedureeval}). 

    \subsubsection{Stabilizing Performance}
    \label{subsec:stabilizingperformance}
\begin{figure}[ht]
        \centering
        \includegraphics[width=1.0\linewidth]{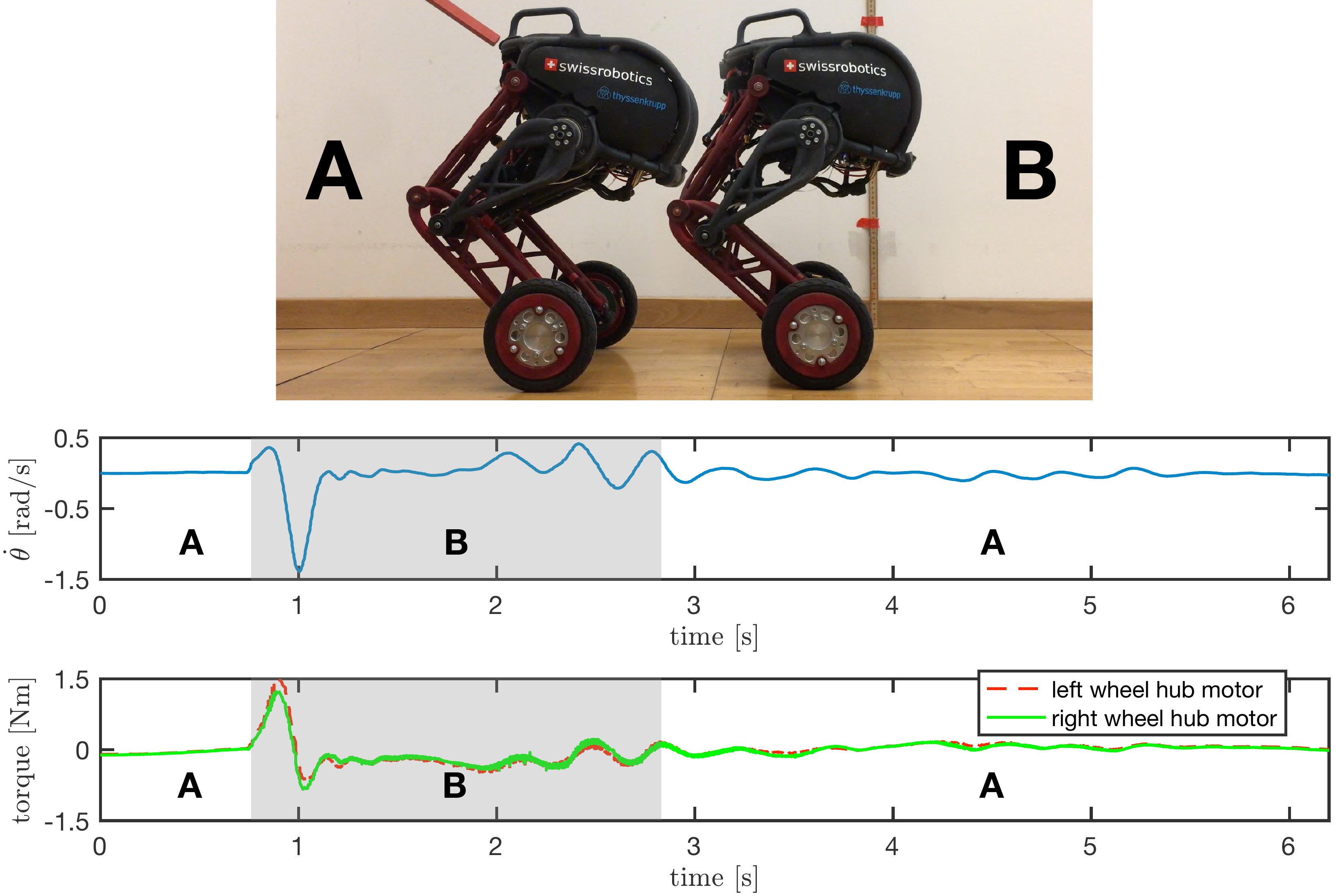}
        \caption{Image: The system on the left (A) is in an unstable equilibrium position and about to be disturbed by a red stick. On the right side (B) the robot is recovering from the disturbance.
        Graphs: The top graph shows the angular velocity $\dot{\theta}$ response, and the bottom one shows the left and right wheel motor torque responses $\vec{u}$ over time.}
        \label{fig:stabilizingperformancegraphs}
    \end{figure}

    The goal of the system is to stay upright at all times.
    Without compromising stability, the system is able to sustain large external disturbances.
    The robot's response to a hit from a wooden stick is shown in \autoref{fig:stabilizingperformancegraphs}.
    Beside impulsive disturbances, the system is also able to go back to its equilibrium position if permanent or longer-lasting disturbances are applied to the system (pulling, pushing or adding weight). 
    
    \subsubsection{Jumping}
    \label{subsec:jumpingevalu}
\begin{figure}[ht]
        \centering
        \includegraphics[width=1.0\linewidth]{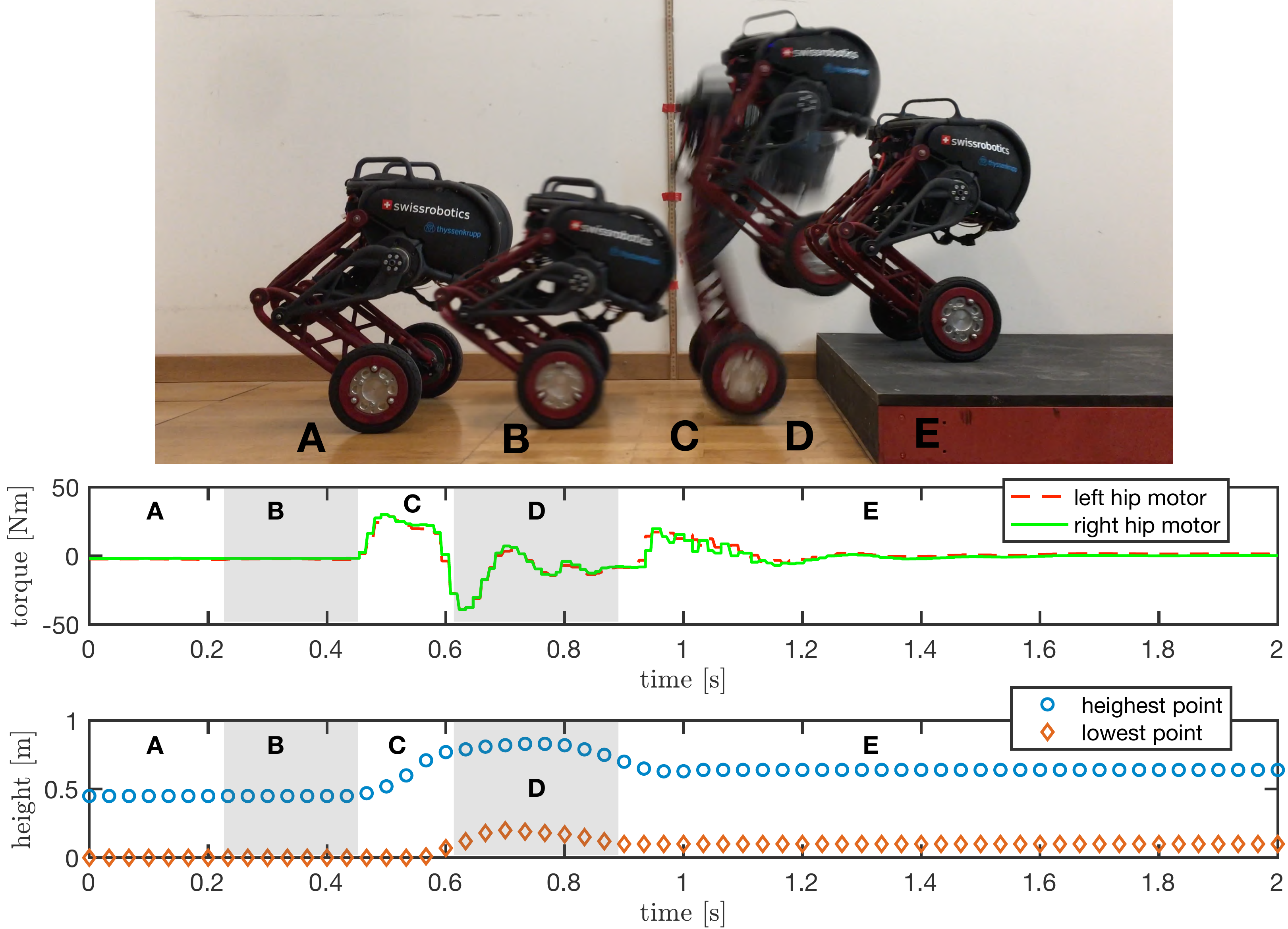}
        \caption{Image: The jump of the system onto a real step with the five phases: Retract Legs (A), Trigger Jump (B), Extract Legs (C), Fly Phase (D) and Land (D). Graphs: In the top graph the hip torques and in the bottom the highest and lowest points of the system are plotted over time during a jump cycle.}
        \label{fig:ascentojumponstep}
    \end{figure}

    The system is able to jump on small steps with a height of \SI{10}{\centi\meter} as demonstrated in \autoref{fig:ascentojumponstep}. 
    The robot needs at least \SI{90}{\centi\meter} both before and after a step to accelerate to the target velocity and in order to land safely.
    
    \subsubsection{Fall Recovery}
    \label{subsec:standupprocedureeval}
\begin{figure}[ht]
        \centering
        \includegraphics[width=1.0\linewidth]{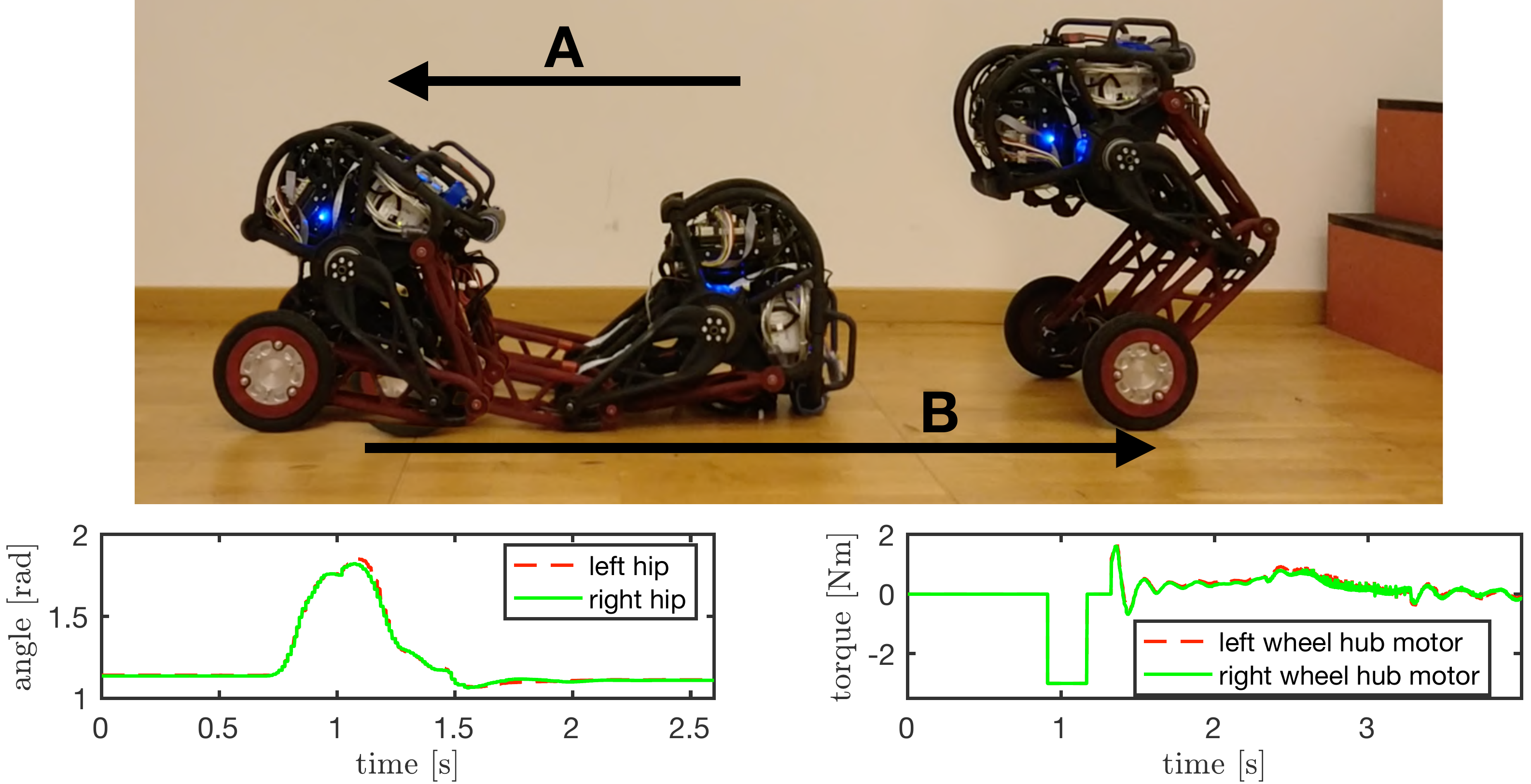}
        \caption{Image: The system is able to stand up by first getting itself into a sitting position (A) and then to the standing position (B). Graphs: The left graph shows the hip motor angle movements in order that the system sits up (A). The right graph represents the wheel motor torque commands for standing up (B).}
        \label{fig:ascentostandup}
    \end{figure}

    The standing up procedure from all recoverable resting positions requires less than \SI{2}{\meter} of space.
    In \autoref{fig:ascentostandup} the stand up procedure is shown for the laying position.
    
%% file: chapters/6.features.tex
Besides the capabilities tested on the real-world prototype, the system also has some experimental features which are to this date only validated in the simulation environment.
An exploration algorithm was developed which can, depending on a local and global planning strategy, map a previously unknown environment.
Using the occupancy information of such a map, the robot is able to find and follow a collision-free path from its current position to any desired destination \cite{Morra2018}, \cite{Kueng2018}.
Additionally, a force-compensating lean mode is introduced to allow the system to be able to turn faster without tipping sideways. 
This is done by varying each leg angle individually to tilt the robot inwards in $\beta$.
The system's unique design and modular setup make it also an interesting object for further academic investigation, e.g., developments of new control strategies such as balancing on one leg \cite{Pfister2018} or \ac{MPC}.

%% file: chapters/7.conclusion.tex
This work presented the \textit{Ascento} platform, a two-wheeled balancing robot that is able to navigate quickly on flat surfaces and to overcome obstacles by jumping.
The system's topology optimized, 3D-printed mechanical design has proven to be both lightweight and impact-resistant.
Additionally, a robust, model-based \ac{LQR} controller has been successfully implemented on the prototype system.
In multiple experiments with the prototype system we demonstrated autonomous jumping onto a step and the ability to recover from downfalls into various positions.
Finally, the robot's operating environment could be extended from indoor to outdoor by incorporating terrain-adaptive principles of legged robotics \cite{gehring2016practice}.